\documentclass[10pt,twocolumn,letterpaper]{article}

\usepackage{iccv}
\usepackage{times}
\usepackage{epsfig}
\usepackage{graphicx}
\usepackage{amsmath}
\usepackage{amssymb}
\usepackage{multirow}
\usepackage{enumitem}

\usepackage{authblk}

\usepackage[pagebackref=true,breaklinks=true,letterpaper=true,colorlinks,bookmarks=false]{hyperref}

\iccvfinalcopy 

\def\iccvPaperID{7569} 

\ificcvfinal\fi

\begin{document}

\title{Degradation-Guided Meta-Restoration Network for Blind Super-Resolution}

\author[1]{\vspace{-1cm}Fuzhi Yang}
\author[2]{Huan Yang}
\author[3]{Yanhong Zeng}
\author[2]{Jianlong Fu}
\author[1]{Hongtao Lu\vspace{-0.2cm}}
\affil[1]{Department of Computer Science and Engineering, Shanghai Jiao Tong University,}
\affil[2]{Microsoft Research, Beijing, P.R. China,}
\affil[3]{School of Data and Computer Science, Sun Yat-sen University, \authorcr \{yfzcopy0702, htlu\}@sjtu.edu.cn, \{huayan, jianf\}@microsoft.com, zengyh7@mail2.sysu.edu.cn}

\maketitle
\ificcvfinal\thispagestyle{empty}\fi

\begin{abstract}
Blind super-resolution (SR) aims to recover high-quality visual textures from a low-resolution (LR) image, which is usually degraded by down-sampling blur kernels and additive noises. 
This task is extremely difficult due to the challenges of complicated image degradations in the real-world. Existing SR approaches either assume a predefined blur kernel or a fixed noise, which limits these approaches in challenging cases. In this paper, we propose a \textbf{D}egradation-guided \textbf{M}eta-restoration network for blind \textbf{S}uper-\textbf{R}esolution (DMSR) that facilitates image restoration for real cases. 
DMSR consists of a degradation extractor and meta-restoration modules. The extractor estimates the degradations in LR inputs and guides the meta-restoration modules to predict restoration parameters for different degradations on-the-fly. DMSR is jointly optimized by a novel degradation consistency loss and reconstruction losses. 
Through such an optimization, DMSR outperforms SOTA by a large margin on three widely-used benchmarks.
A user study including 16 subjects further validates the superiority of DMSR in real-world blind SR tasks.
\end{abstract}


\section{Introduction}
Image super-resolution (SR) is a fundamental computer vision task, which aims to recover high-resolution textures from a degraded low-resolution (LR) image~\cite{IraniImproving}. Recent success has been achieved by deep neural networks in SR tasks, where numerous architectures have been proposed to improve image quality~\cite{dong2015image, dong2016accelerating, lim2017enhanced, niu2020single, zhang2018image}. Such achievements enable SR methods to be applied in practical applications, such as digital zoom algorithm for mobile cameras~\cite{farsiu2005multiframe}, medical imaging~\cite{oktay2016multi} and satellite imaging~\cite{yildirim2012novel}. 


\begin{figure}
\centering
  \includegraphics[width=\linewidth]{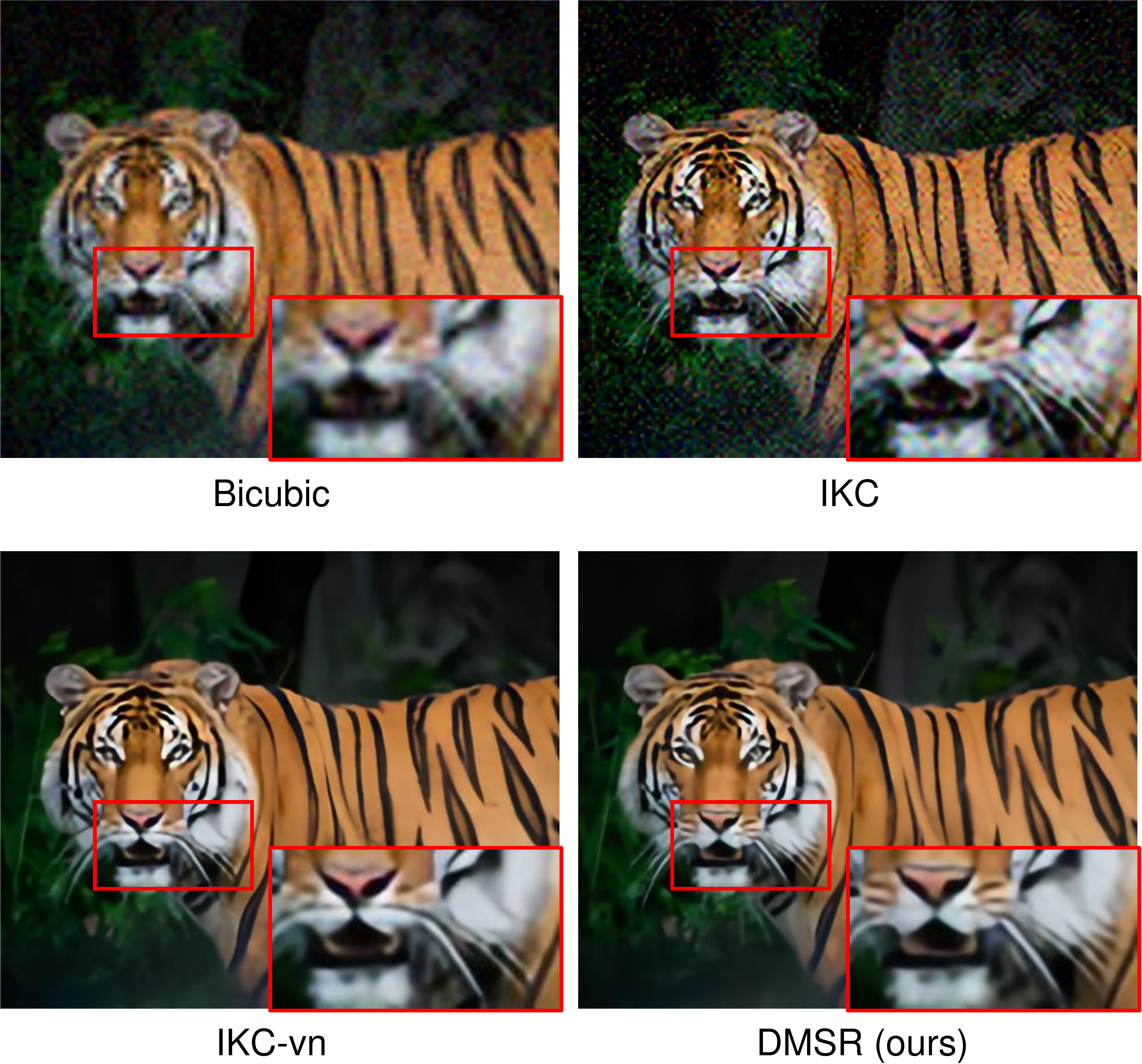}
  \caption{An illustration of real-world blind super-resolution. The LR image is degraded with scale factor $\times 4$, Gaussian blur kernel width 0.2 and Gaussian noise level 15. A SOTA baseline IKC~\cite{gu2019blind} fails as it enlarges the degradation. Even with variant noise (vn) training, IKC-vn can not generate textures as clear as ours.\vspace{-0.5cm}}
  \label{fig:teaser}
\end{figure}

 In general, an image degradation process can be formulated as follows:
\begin{equation}
  \vspace{-0.1cm}I^{LR} = (I^{HR} \otimes k) \downarrow_s + n, \vspace{-0.1cm}
\end{equation}
\label{eq:degradation}
where $\otimes$ indicates convolution operations, and $k$ is the blur kernel. $\downarrow_s$ usually represents bicubic down-sampling, and $n$ indicates additive noises. To generate super-resolved images, most of the existing SR methods either assume a predefined down-sampling blur kernel or a fixed noise level. In particular, earlier works on single image super-resolution (SISR) usually assume a fixed bicubic down-sampling kernel without any blur or noise consideration. Recent progresses have been made by blind super-resolution methods, which aim to super-resolve real-world LR images without degradation knowledge. However, current blind SR methods still have the assumptions of limited degradations. For example, ZSSR~\cite{shocher2018zero} and KernelGAN~\cite{bell2019blind} assume that the degradation is restricted in the LR image and they try to learn internal distribution of the LR image itself. Gu et al.~\cite{gu2019blind} mainly assume the noise level as a fixed value of zero. They propose an iterative corrected kernel estimator IKC to handle different blur degradations. Helou et al.~\cite{elhelou2020stochastic} propose a new module based on existing SR backbones such as~\cite{gu2019blind}, which has the same limitations with previous works. Nonetheless, in real-world scenarios, the degradation process is typically accompanied with variant complicated down-sampling kernels (e.g. Gaussian blur and motion blur), and noises (e.g. Gaussian noise and salt noise). The performance of the SR methods trained on a predefined degradation drops severely when facing different degradation types~\cite{gu2019blind}, thus these methods are still limited in many real-world applications.

To solve the above problems, we propose a \textbf{D}egradation-guided \textbf{M}eta-restoration network for blind \textbf{S}uper-\textbf{R}esolution (DMSR), which effectively super-resolve LR images with arbitrary real-world degradations. 
The blur and noise degradations are estimated on-the-fly and are further used to guide the meta-restoration process in SR networks. 
Specifically, DMSR consists of a degradation extractor and meta-restoration modules. The degradation extractor estimates the blur kernel and the noise map from the degraded LR image. This extractor is optimized by a degradation reconstruction loss. Besides, a novel degradation consistency loss is further proposed to enhance the accuracy of degradation estimation. Such a design plays a key role to guide the meta-restoration modules to generate accurate restoration parameters for different degradations. 
The proposed meta-restoration modules include a \textbf{M}eta-de\textbf{N}oise \textbf{M}odule (MNM) and a \textbf{M}eta-de\textbf{B}lur \textbf{M}odule (MBM) for noise removal and blur recovery, respectively. MNM and MBM take advantage of the estimated degradations and generate meta-biases and meta-weights, respectively. 
Such meta-biases and meta-weights will then serve as the restoration parameters in network to effectively restore high-resolution image features. In summary, to generate super-resolved results, our DMSR model can adjust network parameters according to different degradation situations, such that it can handle different real-world degradations. As shown in Figure~\ref{fig:teaser}, our model is capable of handling images with complicated blur and noise degradations in real-world cenarios.

The main contributions can be summarized as:
\begin{itemize}[nosep]
  \item To the best of our knowledge, we are the first to address variant blur and noise degradations in one end-to-end model for blind SR. We propose two specially-designed meta-restoration modules MBM and MNM to handle blur and noise degradations in the real-world.
  \item We design a degradation extractor in which the degradations are estimated dynamically and are further used to guide the meta-restoration process for SR networks. A degradation consistency loss is further proposed to enhance the estimation accuracy.
  \item Evaluations on three widely-used benchmarks and real images demonstrate that our DMSR model achieves the state-of-the-art performance in blind SR tasks.
\end{itemize}


\section{Related Work}
\textbf{Single image super-resolution}. SISR aims to super-resolve the single LR image to the HR image. Most SISR methods assume the down-sampling blur kernel is predefined (usually bicubic) without any blur and noise consideration. SRCNN~\cite{dong2015image} is the first CNN based SR method while Dong et al.~\cite{dong2016accelerating} further accelerated SR inference process by putting most of the layers in the low-resolution scale. With larger networks or novel optimizations, performance improvement has been achieved in~\cite{kim2016accurate, kim2016deeply, ledig2017photo, lim2017enhanced,  shi2016real, tai2017image, tai2017memnet, tong2017image, wang2018esrgan, zhang2018residual}. 
EDSR~\cite{lim2017enhanced} further improved performance by removing normalization layers in residual blocks. Tong et al.~\cite{tong2017image} and Zhang et al.~\cite{zhang2018residual} used dense blocks~\cite{huang2017densely} for SR task. RCAN~\cite{zhang2018image} adds channel attention in residual blocks as a basic block to achieve the SOTA results. 
Soon afterwards, some methods which consider internal or hierarchical feature correlations were proposed in~\cite{dai2019second, liu2020residual, mei2020image, niu2020single}.
In spite of the above achievements on SISR, these methods are still far from the real-world scenarios as they fail to handle the LR image beyond the assumed degradation.

\begin{figure*}
\centering
  \includegraphics[width=\linewidth]{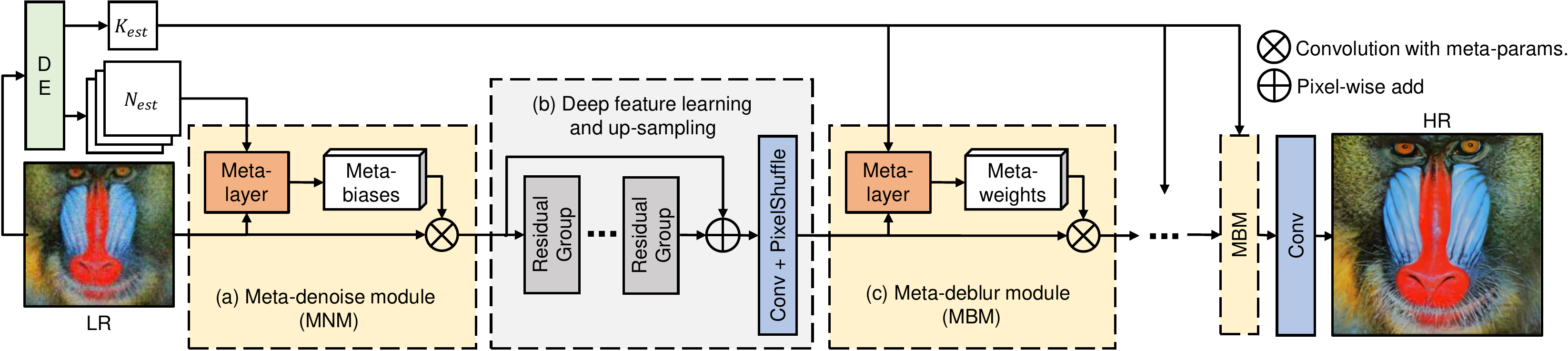}%
  \caption{An overview of our proposed DMSR model. The degradation extractor (DE) estimates the blur kernel $K_{est}$ and the noise map $N_{est}$ from the LR image. In the main SR network: (a) MNM learns to handle noise degradation where the meta-layer generates meta-biases as the convolution parameters according to the estimated noise map $N_{est}$. (b) Several residual groups~\cite{zhang2018image} are adopted to learn the deep features, followed by an up-sampling process. (c) MBM learns to handle blur degradation where the meta-layer outputs the convolution meta-weights guided by the estimated blur kernel $K_{est}$.}
  \label{fig:DMSR}
\end{figure*}

\textbf{Blind super-resolution}. To address real-world super-resolution, some non-blind SR methods which assume degradation parameters are known were firstly proposed~\cite{xu2020unified, zhang2020deep, zhang2018learning, zhang2019deep}. Such a task can be regarded as an intermediate step and upper bound of blind SR, but it is still limited in practical applications as the degradation parameters are unknown in real-world scenarios. Blind SR aims to super-resolve real-world LR images whose degradation information is unavailable. ZSSR~\cite{shocher2018zero} and KernelGAN~\cite{bell2019blind} 
learned the internal distribution of the degraded LR image to construct training pairs by treating LR as the training target. Nonetheless, the degradation assumption was restricted on the LR image itself and the degraded LR image was not an optimal training target in severe degradation situation. KMSR~\cite{zhou2019kernel} used GAN to augment the blur-kernel pool for more SR training pairs. Hussein et al.~\cite{hussein2020correction} proposed a closed-form correction filter that transformed the LR image to adapt to existing leading SISR methods. However, these two methods only addressed blur degradation with a noise-free assumption. Gu et al.~\cite{gu2019blind} proposed an iterative corrected kernel estimator IKC for their non-blind SR method SFTMD, but they mainly focused on blur degradation under the situation of a fixed noise level. Helou et al.~\cite{elhelou2020stochastic} further addressed the overfitting problem from the perspective of frequency domain to improve blind SR performance. However, their module was based on current SR backbones such as Gu et al.~\cite{gu2019blind}, which had the same drawbacks with previous works.

Up till now, most existing SR methods either assume a predefined down-sampling blur kernel or a fixed noise level, which is still far from actual applications. Different from that, we propose a DMSR model which can handle LR images with arbitrary blur and noise degradations.

\section{Approach}
In this section, we present the details of the proposed \textbf{D}egradation-guided \textbf{M}eta-restoration network for blind \textbf{S}uper-\textbf{R}esolution (\textbf{DMSR}). DMSR takes as input a degraded LR image and outputs a restored HR image. 
As shown in Figure \ref{fig:DMSR}, DMSR consists of a degradation extractor and three restoration modules for denoising, upsampling and deblurring, respectively. 
The degradation extractor (DE) estimates the degradations of the LR input. Such a kind of estimated degradations are further leveraged by the meta-restoration modules (MNM and MBM) to restore the image. The full model is optimized by a novel degradation consistency loss and reconstruction losses. We introduce the details of meta-restoration modules in Section~\ref{sec:mrm}. The degradation extractor is 
introduced in Section~\ref{sec:de} and the loss functions are discussed in Section~\ref{sec:loss}.

\subsection{Meta-Restoration Modules}\label{sec:mrm}

As described in E.q.~(\ref{eq:degradation}), an HR image is sequentially degraded by blur, down-sampling and noise. Inspired by this process, we attach the MNM to the head of the network and the MBM to the end of the network, which inversely handles the degradation. 

\textbf{Meta-denoise module (MNM).} There is a meta-layer at the beginning of the meta-denoise module. The guidance for the meta-layer in MNM is the estimated noise map $N_{est}$ from our degradation extractor instead of a noise level $\sigma$ which is commonly used in ~\cite{xu2020unified, zhang2018learning}. Considering the widely-used noise type, Additive White Gaussian Noise (AWGN), the probability density for each pixel $x$ is denoted as:
\begin{align}
    p(x) = \frac{1}{\sigma_n \sqrt{2\pi}}exp(-\frac{x^2}{2\sigma_n^2}),
\end{align}
where $\sigma_n$ is the noise level parameter and $p(x)$ represents the probability density distribution. From this formulation, we can see that the parameter $\sigma_n$ only influences the probability density distribution of the noise in pixel $x$. However, it lacks the ability to express the exact noise value in that pixel. Therefore, we choose the noise map which provides more dense information in our model setting. Ideally, the estimated noise map is the additive noise of the input LR image and the network parameters need to learn an inverse operation. In such a case, the noise map can be regarded as the biases for each spatial position. A concatenation operation followed by a convolution layer are adopted to implement MNM, which can be formulated as:
\begin{align}
     F = {\rm Conv(Concat}(I^{LR}, N_{est})),
\end{align}
where ``Conv'' and ``Concat'' represent the convolution layer and the concatenation operation, respectively. $I^{LR}$ is the input degraded LR image and $N_{est}$ indicates the estimated noise map from our degradation extractor. $F$ is the output feature maps of MNM. When the convolution kernel size is set to 1, each value in the noise map just influences the corresponding position's bias in this convolution process. In other words, the meta-layer in MNM generates meta-biases according to the estimated noise map for the network to better handle noise removal.

\textbf{Deep feature learning and up-sampling.} Between MNM and MBM, there is a network for deep feature learning and up-sampling. We adopt the residual group structure proposed in SOTA SISR method RCAN~\cite{zhang2018image}. Each residual group is composed of several sequential residual channel attention blocks (RCABs) with a long skip connection. The residual path in each RCAB consists of ``convolution + ReLU + convolution + channel attention''. Finally a convolution layer and a pixelshuffle layer enlarge the feature resolution according to the magnification scale factor.

\textbf{Meta-deblur module (MBM).} In general, a Gaussian blur process can be formulated as a convolution process:
\begin{align}
    V = X \otimes K, 
\end{align}
where $X$ and $V$ are input features and blurred features, respectively. $K$ is the blur kernel with the size of $k \times k$ and $\otimes$ represents the convolution operation. Since convolution is a operation on local area, so each pixel value in $V$ is calculated from a local patch with the size of $k \times k$ in $X$. Considering deblur as an inverse process of the above convolution, each piexel value in $X$ is also related to a local patch with the size of $k \times k$ in $V$, which can be also expressed by a convolution process.

In our blind SR network, MBM is to process such an inverse process. To further enhance the expression capability of the module, we make the convolution kernels spatially variant and each kernel can be regarded as a meta-weight which is generated from the meta-layer in MBM. The estimated blur kernel $K_{est}$ serves as a guidance of this meta-layer because the blur kernel influences the deblur process.

Therefore, the target of the meta-layer in MBM is to generate proper meta-weights for this dynamic convolution process. Specifically, the estimated blur kernel will be firstly dimension-reduced to $C_k$ by a fully connected layer. This fully connected layer is initialized by the PCA matrix~\cite{zhang2018learning} which records the principle information estimated from a series of random sampled blur kernels. Then we repeatedly stretch such $1 \times 1 \times C_k$ tensor to the features with $H \times W \times C_k$  where $H$ and $W$ indicate the feature resolution. The stretched features are concatenated with the network features to generate the meta-weights of this dynamic convolution layer. The meta-weights have the shape $k^2 \times H \times W$. Each spatial position with size $k^2 \times 1 \times 1$ represents a convolution kernel in that position. In such a design, the meta-layer in MBM predicts dynamic convolution parameters according to the estimated blur kernel $K_{est}$ for the network to address blur recovery.

\begin{figure}
\centering
  \includegraphics[width=\linewidth]{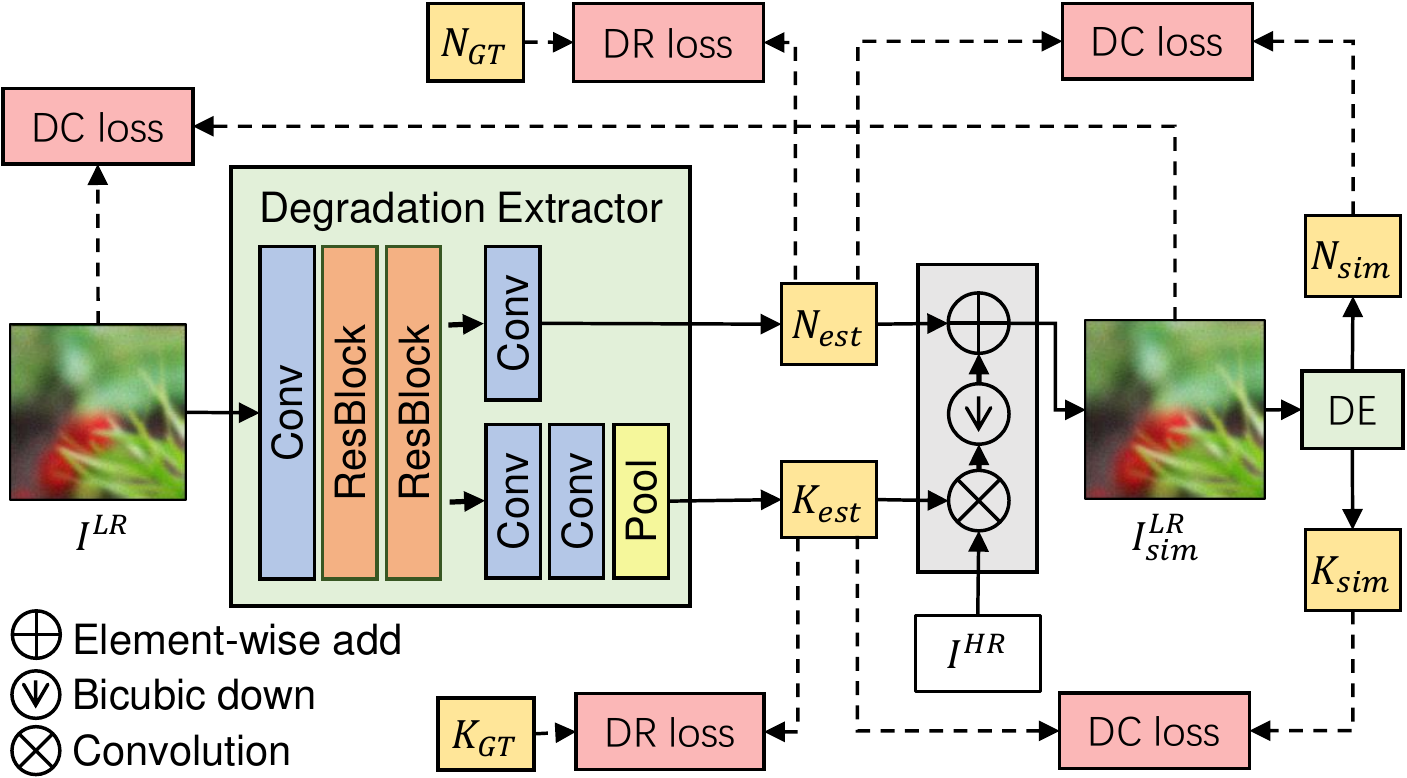}%
  \caption{The proposed degradation extractor estimates a blur kernel and a noise map, which is optimized by a degradation reconstruction loss and a degradation consistency loss. $I^{LR}$ and $I^{LR}_{sim}$ represent the input degraded LR image and the estimated LR image by applying the estimated degradation to the HR image, respectively. $N_{GT}$ and $K_{GT}$ are the ground truth noise map and the ground truth blur kernel. ``DE'' indicates the degradation extractor.}
  \label{fig:de}
\end{figure}

\subsection{Degradation Extractor}\label{sec:de}
It is essential to extract accurate degradation for blind SR task, since degradation mismatch will produce unsatisfactory results~\cite{gu2019blind}. The goal of the degradation extractor is to estimate accurate degradation which can provide solid guidance to the meta-restoration modules (MNM and MBM).

The structure of the degradation extractor is shown in Figure~\ref{fig:de}. One convolution layer followed by two residual blocks are firstly used to extract features of the input LR image. Then there are two branches to extract noise maps and blur kernels respectively. The upper branch estimates the noise map of the input LR image by an additional convolution layer, while the lower branch extracts blur kernel. Two convolution layers and a global average pooling layer are used in the end of the module. The resulting features are reshaped to the size of the blur kernel. We use a softmax layer at the end of the blur branch to ensure there is no value shift before and after blur degradation. To accurately estimate blur kernels and noise maps, we introduce a degradation reconstruction loss and a degradation consistency loss. These two losses are discussed in Section~\ref{sec:loss}.

\subsection{Loss Function}\label{sec:loss}
There are three types of loss functions in our model. The overall loss is denoted as:
\begin{align}
    \mathcal{L}_{overall} = \lambda_{RE} \mathcal{L}_{RE} + \lambda_{DR} \mathcal{L}_{DR} + \lambda_{DC} \mathcal{L}_{DC}, 
\end{align}
where $\mathcal{L}_{RE}$ represents reconstruction loss, which is:
\begin{align}
    \mathcal{L}_{RE} = \frac{1}{CHW} \| I^{HR} - I^{SR} \|_1, 
\end{align}
where ($C, H, W$) is the size of the HR image. We utilize $L_1$ loss which has been demonstrated to produce sharper results compared to $L_2$ loss. 

The last two losses are shown in Figure~\ref{fig:de}. $N_{est}$ and $K_{est}$ are the estimated noise map and blur kernel, while $N_{GT}$ and $K_{GT}$ represent the ground truth noise map and the ground truth blur kernel, respectively. The degradation reconstruction loss can be interpreted as:
\begin{align}
    \mathcal{L}_{DR} = \| N_{GT} - N_{est} \|_2^2 + \| K_{GT} - K_{est} \|_2^2. \label{eq:loss_dr}
\end{align}
The degradation reconstruction loss directly constraints the estimated noise map and blur kernel closer to the accurate ones, which provides a direct supervision. To further enhance the accuracy of degradation estimation, we propose a novel degradation consistency loss. We degrade the original HR image by the estimated degradation to get a simulated degraded LR image $I^{LR}_{sim}$, then we also use the degradation extractor to estimate the blur kernel $K_{sim}$ and noise map $N_{sim}$ from such simulated LR image. The degradation consistency loss can be described as:
\begin{align}
    \mathcal{L}_{DC} = &\|I^{LR}_{sim} - I^{LR}_{est} \|_2^2 + \\ \notag
                      &\| N_{sim} - N_{est} \|_2^2 + \| K_{sim} - K_{est} \|_2^2, \label{eq:loss_dc}
\end{align}
where the first term of degradation consistency loss aims to constraint the consistency between the input LR image and the simulated degraded LR image. Similar $I^{LR}_{sim}$ and $I^{LR}_{est}$ indicates the accuracy of the degradation estimation to some extent. Similarly, the second and third terms in degradation consistency loss constraint the consistency of the estimated noise map and blur kernel, which enhance the accuracy of the degradation estimation.

\section{Experiment}
We conduct experiments on both quantitative and qualitative evaluations. We introduce the implementation details in Section \ref{sec:detail}. Evaluations on benchmarks and real cases are presented in Section \ref{sec:e1} and Section \ref{sec:e2}, followed by ablation study experiments in Section \ref{sec:ab}.

\subsection{Implementation Details}
\label{sec:detail}
\textbf{Training setups}.
To synthesize degraded images for training, we use isotropic blur kernels, bicubic down-sampling and additive white Gaussian noise following the common settings used in previous works~\cite{xu2020unified, zhang2018learning}.
Specifically, the blur kernel size is set to $15 \times 15$, 
and the kernel width is randomly and uniformly sampled from the range of [0.2, 3.0]. 
The noise level $\sigma$ varies in the range of [0, 75]. 
During training, we augment images by random horizontal flipping and rotating $90^{\circ}$, $180^{\circ}$ and $270^{\circ}$. Each mini-batch contains 32 LR patches with size $48 \times 48$. 

We set the channel number of residual blocks in the degradation extractor as 64 and the kernel size of all the convolution layers as $3 \times 3$.
There are 5 residual groups with each containing 20 RCABs in our full model.
The global average pooling is used at the end of the blur branch. 
The weight coefficients for $\mathcal{L}_{RE}$, $\mathcal{L}_{DR}$ and $\mathcal{L}_{DC}$ are 1, 10 and 1, respectively. Adam optimizer with $\beta_1 = 0.9$, $\beta_2 = 0.999$ and $\epsilon = 1 \times 10^{-8}$  is used with initial learning rate of $1 \times 10^{-4}$. We train the model for $5 \times 10^5$ iterations and the learning rate is halved every $2 \times 10^5$ iterations.

\textbf{Datasets and Metrics.}
For fair comparisons, we use DF2K~\cite{agustsson2017ntire,timofte2017ntire} as training set following the common settings used in previous works \cite{gu2019blind, xu2020unified}. 
All models are evaluated on both standard benchmarks (i.e., Set5~\cite{bevilacqua2012low}, Set14~\cite{zeyde2010single} and B100~\cite{martin2001database}) and real-world cases. Specifically, the real-world cases we used for comparisons include the commonly-used real image \textit{Flower}~\cite{lebrun2015noise} and the test set of ``NTIRE 2020 Real World Super-Resolution'' challenge~\cite{lugmayr2020ntire}. The test set of the challenge contains 100 unknown-degraded test images without ground truth. 

We report quantitative results in terms of PSNR and SSIM metrics, which are calculated on Y channel of YCbCr space. Since the ground truths of the degraded images of real cases are unavailable, we conduct qualitative evaluations and a user study for fair comparisons. 


\begin{table*}
\caption{Performance comparison among different blind SR methods (PSNR / SSIM). $\sigma_k$ and $\sigma_n$ are blur kernel width and AWGN level, respectively. ``+'' indicates connection of a denoise method and a blind SR method. Red / blue colors indicate the best and the second best results. [Best viewed in color]}
\label{tab:quantitative}
\centering
\scalebox{0.7}{
\begin{tabular}{lcccccccccc}
\hline
\multirow{2}{*}{Methods} &[$\sigma_k$, &\multicolumn{3}{c}{Set5~\cite{bevilacqua2012low}} & \multicolumn{3}{c}{Set14~\cite{zeyde2010single}} & \multicolumn{3}{c}{B100~\cite{martin2001database}} \\
 &$\sigma_n$] &$\times$2 &$\times$3 &$\times$4 &$\times$2 &$\times$3 &$\times$4 &$\times$2 &$\times$3 &$\times$4 \\
\hline

\hline
ZSSR~\cite{shocher2018zero} & &26.60 / .5972 &25.62 / .5836 &24.58 / .5600 &25.73 / .5893 &24.48 / .5382 &23.62 / .5091 &25.31 / .5667 &24.19 / .5079 &23.43 / .4724 \\
IKC~\cite{gu2019blind} & &26.74 / .7582 &26.96 / .6546 &23.73 / .5357 &25.25 / .6732 &25.36 / .5943 &21.47 / .4312 &25.38 / .6404 &24.93 / .5540 &21.16 / .3852 \\
SFM~\cite{elhelou2020stochastic} &[0.2, &28.13 / .6856 &24.60 / .5502 &20.80 / .4333 &26.81 / .6561 &23.76 / .5213 &18.42 / .3251 &26.47 / .6315 &23.33 / .4802 &18.46 / .2924 \\
IKC~\cite{gu2019blind}-vn &15] &{\color{blue}{32.54}} / {\color{blue}{.8980}} &29.84 / .8551 &{\color{blue}{28.79}} / {\color{blue}{.8261}} &{\color{blue}{30.04}} / {\color{blue}{.8382}} &27.76 / .7618 &26.54 / {\color{red}{.7122}} &{\color{blue}{29.17}} / {\color{blue}{.8066}} &{\color{blue}{26.97}} / {\color{red}{.7221}} &{\color{blue}{25.93}} / {\color{red}{.6629}} \\
SFM~\cite{elhelou2020stochastic}-vn & &32.30 / .8962 &{\color{blue}{30.26}} / {\color{blue}{.8595}} &28.63 / .8232 &29.68 / .8350 &{\color{blue}{27.87}} / {\color{blue}{.7621}} &{\color{blue}{26.57}} / .7102 &28.89 / .8054 &26.93 / .7182 &25.92 / .6608 \\
DMSR (ours) & &{\color{red}{32.79}} / {\color{red}{.9009}} &{\color{red}{30.64}} / {\color{red}{.8637}} &{\color{red}{29.06}} / {\color{red}{.8285}} &{\color{red}{30.30}} / {\color{red}{.8417}} &{\color{red}{28.12}} / {\color{red}{.7662}} &{\color{red}{26.74}} / {\color{red}{.7122}} &{\color{red}{29.33}} / {\color{red}{.8109}} &{\color{red}{27.23}} / {\color{blue}{.7197}} &{\color{red}{26.05}} / {\color{blue}{.6626}} \\

\hline
ZSSR~\cite{shocher2018zero} & &25.33 / .5346 &25.06 / .5478 &24.28 / .5374 &24.29 / .4891 &23.95 / .4909 &23.40 / .4833 &24.06 / .4546 &23.77 / .4554 &23.28 / .4469\\
IKC~\cite{gu2019blind} & &25.69 / .7128 &25.77 / .6085 &23.46 / .5038 &24.25 / .6152 &24.47 / .5400 &21.82 / .4111 &24.54 / .5827 &24.26 / .4980 & 20.60 / .3343 \\
SFM~\cite{elhelou2020stochastic} &[1.3, &26.15 / .6168 &24.59 / .5233 &21.64 / .4400 &24.84 / .5523 &23.53 / .4738 &19.23 / .3243 &24.78 / .5187 &23.27 / .4337 &19.39 / .2933 \\
IKC~\cite{gu2019blind}-vn &15] &30.56 / .8602 &29.13 / .8278 &{\color{blue}{28.26}} / {\color{blue}{.8072}} &28.15 / .7654 &26.97 / .7195 &{\color{blue}{26.24}} / .6881 &27.34 / .7178 &26.34 / .6705 &{\color{blue}{25.65}} / .6381 \\
SFM~\cite{elhelou2020stochastic}-vn & &{\color{blue}{30.66}} / {\color{blue}{.8622}} &{\color{blue}{29.15}} / {\color{blue}{.8304}} &28.00 / .8016 &{\color{blue}{28.27}} / {\color{blue}{.7717}} &{\color{blue}{27.11}} / {\color{blue}{.7242}} &26.19 / {\color{blue}{.6885}} &{\color{blue}{27.45}} / {\color{red}{.7260}} &{\color{blue}{26.42}} / {\color{red}{.6780}} &25.64 / {\color{blue}{.6384}} \\
DMSR (ours) & &{\color{red}{31.13}} / {\color{red}{.8698}} &{\color{red}{29.65}} / {\color{red}{.8403}} &{\color{red}{28.48}} / {\color{red}{.8116}} &{\color{red}{28.51}} / {\color{red}{.7735}} &{\color{red}{27.24}} / {\color{red}{.7261}} &{\color{red}{26.29}} / {\color{red}{.6895}} &{\color{red}{27.54}} / {\color{blue}{.7252}} &{\color{red}{26.46}} / {\color{blue}{.6738}} &{\color{red}{25.70}} / {\color{red}{.6389}}\\

\hline
ZSSR~\cite{shocher2018zero} & &23.42 / .4378 &23.40 / .4643 &23.26 / .4826 &22.53 / .3761 &22.50 / .3989 &22.43 / .4202 &22.63 / .3436 &22.64 / .3683 &22.54 / .3888\\
IKC~\cite{gu2019blind} & &24.33 / .6511 &23.83 / .5079 &21.60 / .3993 &23.21 / .5606 &22.85 / .4443 &20.15 / .3058 &23.71 / .5334 &23.01 / .4111 &19.58 / .2569 \\
SFM~\cite{elhelou2020stochastic} &[2.6, &24.13 / .5255 &23.17 / .4442 &19.67 / .3246 &23.07 / .4523 &22.29 / .3820 &18.27 / .2453 &23.35 / .4225 &22.28 / .3458 &18.51 / .2192 \\
IKC~\cite{gu2019blind}-vn &15] &{\color{blue}{27.57}} / {\color{blue}{.7849}} &26.52 / .7523 &{\color{blue}{26.38}} / {\color{blue}{.7503}} &{\color{blue}{25.52}} / {\color{blue}{.6645}} &24.86 / .6373 &{\color{blue}{24.79}} / {\color{blue}{.6318}} &{\color{blue}{25.25}} / {\color{blue}{.6182}} &24.74 / .5935 &{\color{blue}{24.64}} / {\color{blue}{.5875}} \\
SFM~\cite{elhelou2020stochastic}-vn & &27.08 / .7686 &{\color{blue}{26.80}} / {\color{blue}{.7602}} &26.18 / .7425 &25.38 / .6578 &{\color{blue}{25.02}} / {\color{blue}{.6423}} &24.71 / .6287 &25.19 / .6141 &{\color{blue}{24.94}} / {\color{blue}{.6040}} &24.60 / .5855 \\
DMSR (ours) & &{\color{red}{28.81}} / {\color{red}{.8171}} &{\color{red}{27.79}} / {\color{red}{.7943}} &{\color{red}{27.10}} / {\color{red}{.7743}} &{\color{red}{26.43}} / {\color{red}{.6916}} &{\color{red}{25.81}} / {\color{red}{.6672}} &{\color{red}{25.30}} / {\color{red}{.6478}} &{\color{red}{25.86}} / {\color{red}{.6428}} &{\color{red}{25.33}} / {\color{red}{.6173}} &{\color{red}{24.93}} / {\color{red}{.6005}}\\

\hline
ZSSR~\cite{shocher2018zero} & &17.33 / .2164 &17.06 / .2142 &16.89 / .2265 &17.09 / .2167 &16.75 / .1982 &16.54 / .1936 &16.95 / .2011 &16.63 / .1777 &16.43 / .1700\\
IKC~\cite{gu2019blind} & &23.41 / .5976 &17.49 / .2372 &14.54 / .1766 &22.71 / .5137 &17.21 / .2191 &12.13 / .1013 &23.14 / .4876 &16.74 / .1898 &11.77 / .0811 \\
SFM~\cite{elhelou2020stochastic} &[0.2, &18.36 / .2468 &15.52 / .1873 &11.93 / .1112 &17.66 / .2346 &15.31 / .1723 &11.88 / .0955 &17.69 / .2227 &15.35 / .1563 &11.58 / .0812 \\
IKC~\cite{gu2019blind}-vn &50] &{\color{blue}{28.35}} / {\color{blue}{.8176}} &26.28 / .7678 &{\color{blue}{25.12}} / {\color{blue}{.7323}} &{\color{blue}{26.75}} / {\color{blue}{.7278}} &{\color{blue}{25.10}} / {\color{blue}{.6633}} &{\color{blue}{23.99}} / .6195 &{\color{blue}{26.06}} / {\color{blue}{.6763}} &{\color{blue}{24.67}} / {\color{blue}{.6138}} &{\color{blue}{23.81}} / {\color{blue}{.5754}} \\
SFM~\cite{elhelou2020stochastic}-vn & &28.25 / .8165 &{\color{blue}{26.42}} / {\color{blue}{.7715}} &25.10 / .7317 &26.44 / .7251 &25.02 / .6614 &23.98 / {\color{blue}{.6196}} &25.77 / .6753 &24.55 / .6125 &23.78 / .5752 \\
DMSR (ours) & &{\color{red}{28.58}} / {\color{red}{.8226}} &{\color{red}{26.66}} / {\color{red}{.7763}} &{\color{red}{25.38}} / {\color{red}{.7417}} &{\color{red}{26.93}} / {\color{red}{.7296}} &{\color{red}{25.30}} / {\color{red}{.6659}} &{\color{red}{24.21}} / {\color{red}{.6232}} &{\color{red}{26.20}} / {\color{red}{.6796}} &{\color{red}{24.83}} / {\color{red}{.6158}} &{\color{red}{23.94}} / {\color{red}{.5779}}\\

\hline
ZSSR~\cite{shocher2018zero} & &17.12 / .1814 &16.88 / .1906 &16.71 / .2034 &16.84 / .1641 &16.65 / .1697 &16.46 / .1764 &16.70 / .1446 &16.54 / .1483 &16.40 / .1548\\
IKC~\cite{gu2019blind} & &23.07 / .5762 &17.20 / .2094 &13.37 / .1298 &22.46 / .4969 &17.01 / .1880 &12.52 / .0991 &22.91 / .4709 &16.68 / .1611 &11.89 / .0758 \\
SFM~\cite{elhelou2020stochastic} &[1.3, &18.59 / .2243 &15.42 / .1665 &11.88 / .0991 &17.51 / .1817 &15.24 / .1475 &11.87 / .0857 &17.56 / .1654 &15.29 / .1307 &11.59 / .0713 \\
IKC~\cite{gu2019blind}-vn &50] &27.27 / .7862 &{\color{blue}{25.87}} / .7466 &{\color{blue}{24.81}} / {\color{blue}{.7171}} &{\color{blue}{25.73}} / {\color{blue}{.6766}} &{\color{blue}{24.65}} / .6345 &{\color{blue}{23.80}} / .6071 &{\color{blue}{25.24}} / .6267 &{\color{blue}{24.36}} / .5883 &{\color{blue}{23.67}} / {\color{blue}{.5633}} \\
SFM~\cite{elhelou2020stochastic}-vn & &{\color{blue}{27.32}} / {\color{blue}{.7877}} &25.82 / {\color{blue}{.7472}} &24.78 / .7155 &25.69 / .6762 &24.64 / {\color{blue}{.6361}} &23.79 / {\color{blue}{.6074}} &25.23 / {\color{red}{.6281}} &24.32 / {\color{red}{.5904}} &23.66 / {\color{blue}{.5633}} \\
DMSR (ours) & &{\color{red}{27.62}} / {\color{red}{.7942}} &{\color{red}{25.96}} / {\color{red}{.7533}} &{\color{red}{24.98}} / {\color{red}{.7256}} &{\color{red}{25.86}} / {\color{red}{.6776}} &{\color{red}{24.76}} / {\color{red}{.6369}} &{\color{red}{23.90}} / {\color{red}{.6088}} &{\color{red}{25.30}} / {\color{blue}{.6279}} &{\color{red}{24.42}} / {\color{blue}{.5897}} &{\color{red}{23.71}} / {\color{red}{.5643}}\\

\hline
ZSSR~\cite{shocher2018zero} & &16.67 / .1354 &16.61 / .1525 &16.44 / .1693 &16.40 / .1127 &16.29 / .1252 &16.20 / .1433 &16.36 / .0976 &16.28 / .1103 &16.22 / .1276\\
IKC~\cite{gu2019blind} & &22.42 / .5433 &16.62 / .1616 &14.26 / .1412 &21.91 / .4660 &16.66 / .1428 &11.72 / .0608 &22.48 / .4434 &16.40 / .1209 &11.75 / .0590 \\
SFM~\cite{elhelou2020stochastic} &[2.6, &18.05 / .1753 &15.12 / .1255 &11.73 / .0752 &17.25 / .1336 &14.84 / .1023 &11.72 / .0627 &17.28 / .1185 &15.04 / .0937 &11.51 / .0532 \\
IKC~\cite{gu2019blind}-vn &50] &{\color{blue}{25.59}} / {\color{blue}{.7272}} &24.57 / .6925 &{\color{blue}{23.85}} / {\color{blue}{.6763}} &{\color{blue}{24.25}} / {\color{blue}{.6135}} &23.49 / .5875 &{\color{blue}{23.07}} / {\color{blue}{.5742}} &{\color{blue}{24.17}} / {\color{blue}{.5703}} &23.56 / .5484 &{\color{blue}{23.20}} / {\color{blue}{.5371}} \\
SFM~\cite{elhelou2020stochastic}-vn & &25.44 / .7212 &{\color{blue}{24.72}} / {\color{blue}{.6995}} &23.77 / .6712 &24.20 / .6113 &{\color{blue}{23.57}} / {\color{blue}{.5905}} &23.06 / .5733 &24.16 / .5693 &{\color{blue}{23.65}} / {\color{blue}{.5525}} &{\color{blue}{23.20}} / .5370 \\
DMSR (ours) & &{\color{red}{26.05}} / {\color{red}{.7467}} &{\color{red}{24.86}} / {\color{red}{.7139}} &{\color{red}{24.21}} / {\color{red}{.6967}} &{\color{red}{24.56}} / {\color{red}{.6240}} &{\color{red}{23.83}} / {\color{red}{.5999}} &{\color{red}{23.23}} / {\color{red}{.5818}} &{\color{red}{24.36}} / {\color{red}{.5788}} &{\color{red}{23.75}} / {\color{red}{.5566}} &{\color{red}{23.27}} / {\color{red}{.5422}}\\

\hline
\hline 
ZSSR~\cite{shocher2018zero} & &36.98 / .9567 &32.23 / .8982 &29.40 / .8264 &32.77 / .9101 &29.12 / .8206 &27.16 / .7451 &31.44 / .8910 &28.26 / .7858 &26.66 / .7063 \\
IKC~\cite{gu2019blind} &[0.2, &37.26 / .9572 &{\color{blue}{34.02}} / {\color{blue}{.9261}} &31.55 / .8931 &33.06 / .9116 &30.09 / .8421 &28.20 / .7830 &{\color{blue}{31.94}} / {\color{blue}{.8933}} &28.89 / {\color{blue}{.8066}} &27.43 / {\color{blue}{.7376}} \\
SFM~\cite{elhelou2020stochastic} &0] &{\color{blue}{37.53}} / {\color{blue}{.9581}} &33.45 / .9249 &{\color{blue}{32.11}} / {\color{blue}{.8950}} &{\color{blue}{33.23}} / {\color{blue}{.9129}} &{\color{blue}{30.12}} / {\color{blue}{.8424}} &{\color{blue}{28.48}} / {\color{blue}{.7833}} &31.93 / .8928 &{\color{blue}{28.92}} / .8061 &{\color{blue}{27.44}} / .7378 \\
DMSR-nf (ours) & &{\color{red}{37.96}} / {\color{red}{.9617}} &{\color{red}{34.49}} / {\color{red}{.9289}} &{\color{red}{32.37}} / {\color{red}{.8977}} &{\color{red}{33.48}} / {\color{red}{.9174}} &{\color{red}{30.34}} / {\color{red}{.8448}} &{\color{red}{28.73}} / {\color{red}{.7860}} &{\color{red}{32.14}} / {\color{red}{.8995}} &{\color{red}{29.14}} / {\color{red}{.8087}} &{\color{red}{27.65}} / {\color{red}{.7398}}\\

\hline
\end{tabular}
}
\end{table*}

\begin{figure*}
\centering
  \begin{tabular}{p{4mm}p{21mm}p{20mm}p{18mm}p{19mm}p{19mm}p{24mm}p{15mm}p{10mm}}
        \ 
		&\scriptsize{ZSSR~\cite{shocher2018zero}}
		&\scriptsize{IKC~\cite{gu2019blind}}
		&\scriptsize{SFM~\cite{elhelou2020stochastic}}
		&\scriptsize{IKC~\cite{gu2019blind}-vn}
		&\scriptsize{SFM~\cite{elhelou2020stochastic}-vn}
		&\scriptsize{DMSR (ours)}
		&\scriptsize{GT}
		&\ \ \ 
  \end{tabular}\\
  \includegraphics[width=0.95\linewidth]{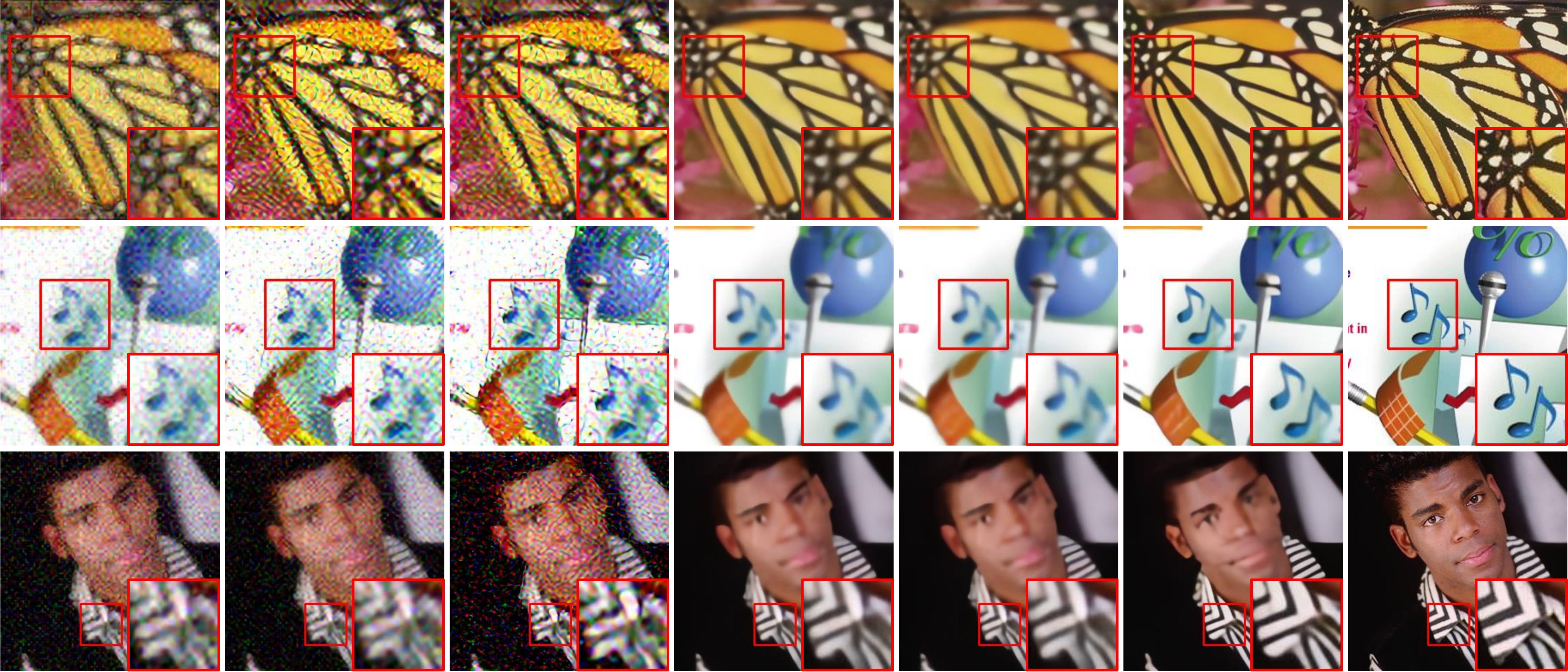}%
  \caption{Visual comparison among different blind SR methods on Set5~\cite{bevilacqua2012low}, Set14~\cite{zeyde2010single} and B100~\cite{martin2001database} datasets sequentially. Degradation: isotropic Gaussian blur kernel 2.6, additive white Gaussian noise 15 and scale factor $\times$4. [Best viewed in color]}
  \label{fig:visual}
\end{figure*}

\subsection{Evaluations on Benchmarks}
\label{sec:e1}
In this section, we compare our model with SOTA blind SR methods, i.e., ZSSR~\cite{shocher2018zero}, IKC~\cite{gu2019blind} and SFM~\cite{elhelou2020stochastic} on standard benchmarks. ZSSR is the SOTA method to learn internal distribution of the LR image. IKC is the SOTA blind SR method and SFM achieves performance improvement based on IKC backbone. For IKC and SFM which assume a fixed noise level, we additionally train IKC-vn and SFM-vn models with variant noise training for more fair comparison. We attach all the websites of the implementation in the footnote\footnote{Implementation of the methods in blind SR comparison:\\
\scriptsize{
ZSSR~\cite{shocher2018zero}: \url{https://github.com/assafshocher/ZSSR}\\
IKC~\cite{gu2019blind}: \url{https://github.com/yuanjunchai/IKC}\\
SFM~\cite{elhelou2020stochastic}: \url{https://github.com/majedelhelou/SFM}\\}.}.
For fair comparisons, we train IKC and SFM with $15 \times 15$ blur kernels on DF2K dataset. We use IKC as the backbone of SFM and the percent rate of training for SFM is 50\%. We sample isotropic Gaussian blur kernels from \{0.2, 1.3, 2.6\} and AWGN levels from \{15, 50\}.

\textbf{Quantitative Comparisons.}
In Table~\ref{tab:quantitative}, ZSSR which learns the internal distribution of the LR image performs badly since the noise and blur degradation in LR images will make its performance drop significantly. For IKC and SFM which have fixed noise level assumption zero, the performance drops significantly when noise level becomes larger. Even with variant noise training, the performance of SOTA methods is increased but still limited since they do not have specially-designed modules to handle variant noises. Among these blind SR methods, our DMSR can achieve the best performance on all benchmarks in different degradation situations.

To further validate the superior performance of our model, we also conduct experiments in the noise-free situation to meet the assumption of IKC and SFM, where IKC and SFM have significantly better performance. To get the noise-free version of our model ``DMSR-nf'', we fine-tune the DMSR where we fix the noise level to 0. We train another $1 \times 10^{5}$ iterations with learning rate $1 \times 10^{-4}$. As shown in the bottom row in Table~\ref{tab:quantitative}, even in this narrow range of degradations, our model can still achieve the best performance. This further verifies the real-world super-resolution capability of our model.

\textbf{Qualitative Comparisons.} We also show visual comparison results among different blind SR methods in Figure~\ref{fig:visual}, where degradation parameters are set as isotropic blur kernel 2.6, noise level $\sigma$ 15 with scale factor $\times$4. ZSSR learns the noise distribution of the LR image so its SR results obviously contain noise. IKC and SFM fails to handle such degraded images beyond their degradation assumption. With variant noise training, IKC-vn and SFM-vn can generate better SR results, but not clear as ours. Our DMSR model can handle both noise and blur degradation well and achieves the best visual performance.


\subsection{Evaluations on Real Cases}
\label{sec:e2}
In this section, we further evaluate different methods on real cases. 
We compare our DMSR model with different kinds of SR methods to show the overall practical application ability. These methods include two SOTA blind SR methods with variant noise training IKC~\cite{gu2019blind}-vn and SFM~\cite{elhelou2020stochastic}-vn. We also adopt a SOTA SISR method RCAN~\cite{zhang2018image} and a SOTA non-blind SR method UDVD~\cite{xu2020unified} to conduct comparison. For the non-blind SR method UDVD, manual grid search on degradation parameters are usually performed and the best result is chosen. However, such a design is time-consuming on real-world application and not fair for other methods since they manually choose the best result from all the generated results. So for fair comparison, we draw a degradation parameter window which contains different degradation types applied on one image. It has 24 cases of degradations with 6 different noise levels $\sigma_n$ and 4 different blur kernels $\sigma_k$ for each scale factor 2, 3, 4, where $\times$4 is shown in Supplementary. During inference, we first manually choose the most similar degraded image in the window and adopt its parameters as the input for UDVD. Such a design can save more real-world inference time and is more fair for comparison.

\begin{figure}
\centering
  \includegraphics[width=\linewidth, page=1]{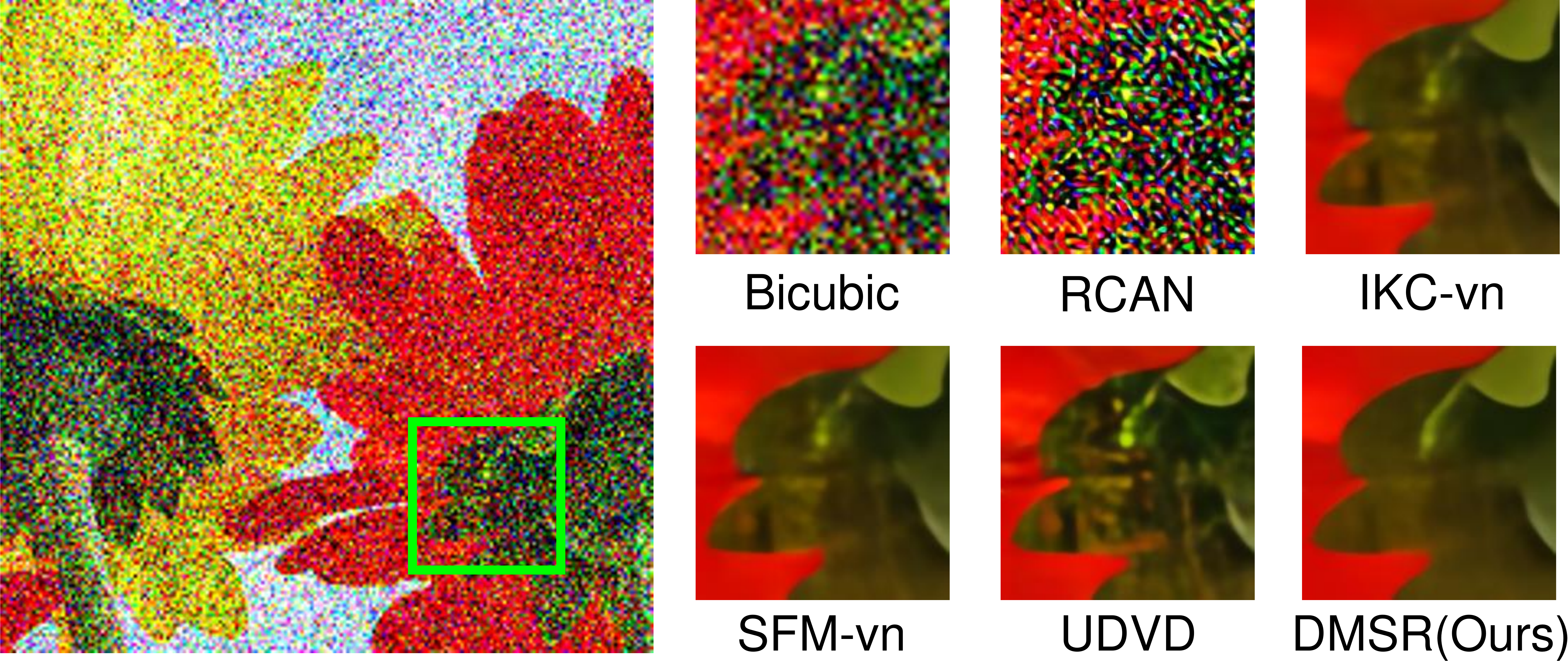}
  \caption{Performance on the real image \textit{Flowers}~\cite{zhang2018learning} among different kinds of SR methods on scale factor $\times$4. We select the parameter [$\sigma_k$=2.1, $\sigma_n$=60] from the degradation parameter window in supplementary for UDVD~\cite{xu2020unified}. [Best viewed in color]}
  \label{fig:visual_real}
\end{figure}

\textbf{Performance on real images}. We first conduct evaluation on the real image \textit{Flowers}~\cite{lebrun2015noise}. Since there is no ground truth HR image for this image, we only show the visual comparison results in Figure~\ref{fig:visual_real}. The degradation parameters [$\sigma_k$, $\sigma_n$] of UDVD is selected as [2.1, 60] for this image. As we can see, our DMSR model can achieve the best visual performance among these methods. The visual quality of RCAN is severely influenced by the noise degradation. IKC-vn and SFM-vn fail to remove the influences of the noise. Even with manually chosen degraded parameters, the SOTA non-blind SR method UDVD cannot generate textures as natural as ours and there are some artifacts in their result.

\begin{figure}
\centering
  \includegraphics[width=\linewidth]{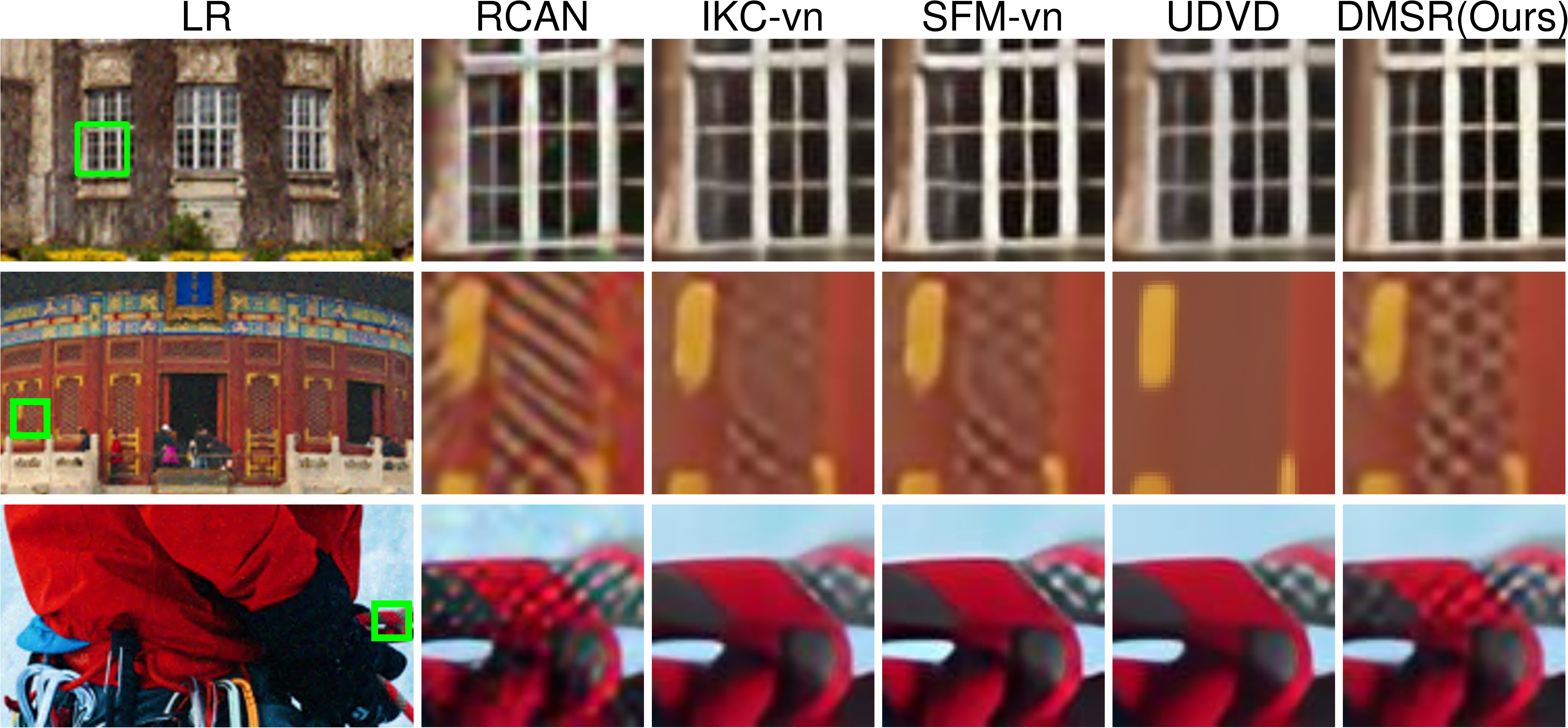}
  \caption{Visual comparison on the test set of ``NTIRE 2020 Real World Super-Resolution'' challenge~\cite{lugmayr2020ntire}. Degradation parameters are set [$\sigma_k$=1.2, $\sigma_n$=15] from the degradation parameter window in supplementary for UDVD~\cite{xu2020unified}. [Best viewed in color]}
  \label{fig:visual_challenge}
\end{figure}

\begin{figure}
\centering
  \includegraphics[width=\linewidth]{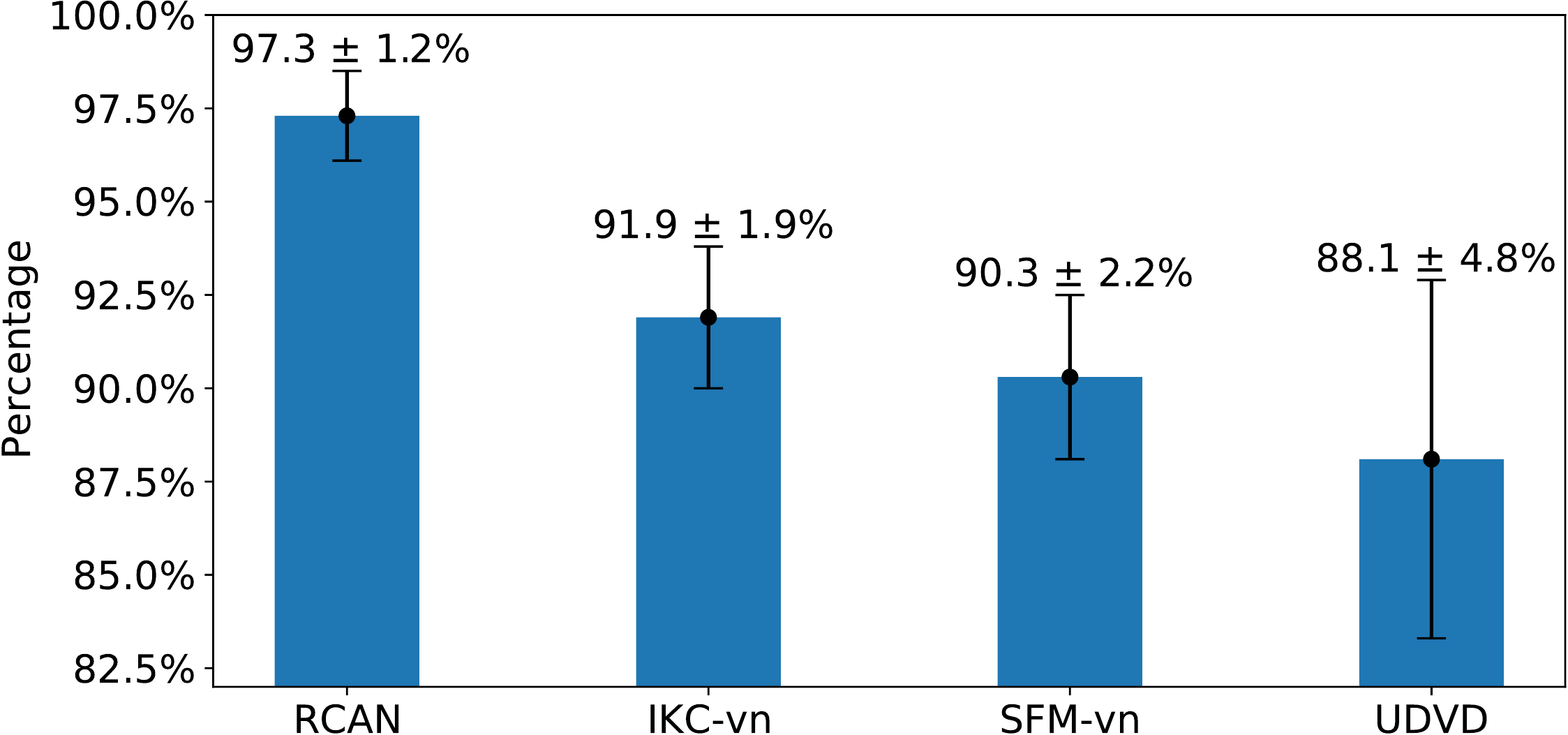}%
  \caption{User study results where values on Y-axis represent the percentage of users that prefer our DMSR over other methods.}
  \label{fig:user_study}
\end{figure}

\textbf{Performance on real-world image challenge}. In addition to the real image, we also run our model on the test set of ``NTIRE 2020 Real World Super-Resolution'' challenge~\cite{lugmayr2020ntire}. There are 100 unknown-degraded test images in Track 1 and we use these images for evaluation. The degradation parameters of UDVD for this test set are set to [1.2, 15]. Visual comparison is shown in Figure~\ref{fig:visual_challenge}. As shown in this figure, our DMSR model can generate the results with clearer and more realistic textures. To further validate the superiority of our model, we conduct a user study on this challenge test set where our DMSR is compared with RCAN, IKC-vn, SFM-vn and UDVD. We collect 3,200 votes from 16 subjects, where each subject is invited to compare our model with two other methods. Therefore, each of the four comparison combination is evaluated by 8 subjects. In each comparison process, the users are provided with two images, including a DMSR result and another method's result. Users are asked to select the one with better visual quality. The user study results are shown in Figure~\ref{fig:user_study}, where the values on Y-axis indicate the percentage of users that prefer our DMSR model over other methods. As we can see, DMSR significantly outperforms SOTA SISR method with over 97\% of users voting for ours. For SOTA blind and non-blind SR methods IKC-vn, SFM-vn and UDVD, our model still has over 88\% probability of winning. Such results validate the favorable visual quality of our DMSR model. This demonstrates that the meta-restoration modules in our model can generate proper network parameters on-the-fly for different real-world degradation situations.

\subsection{Ablation Study}
\label{sec:ab}
\begin{figure}
\centering
  \includegraphics[width=0.66\linewidth]{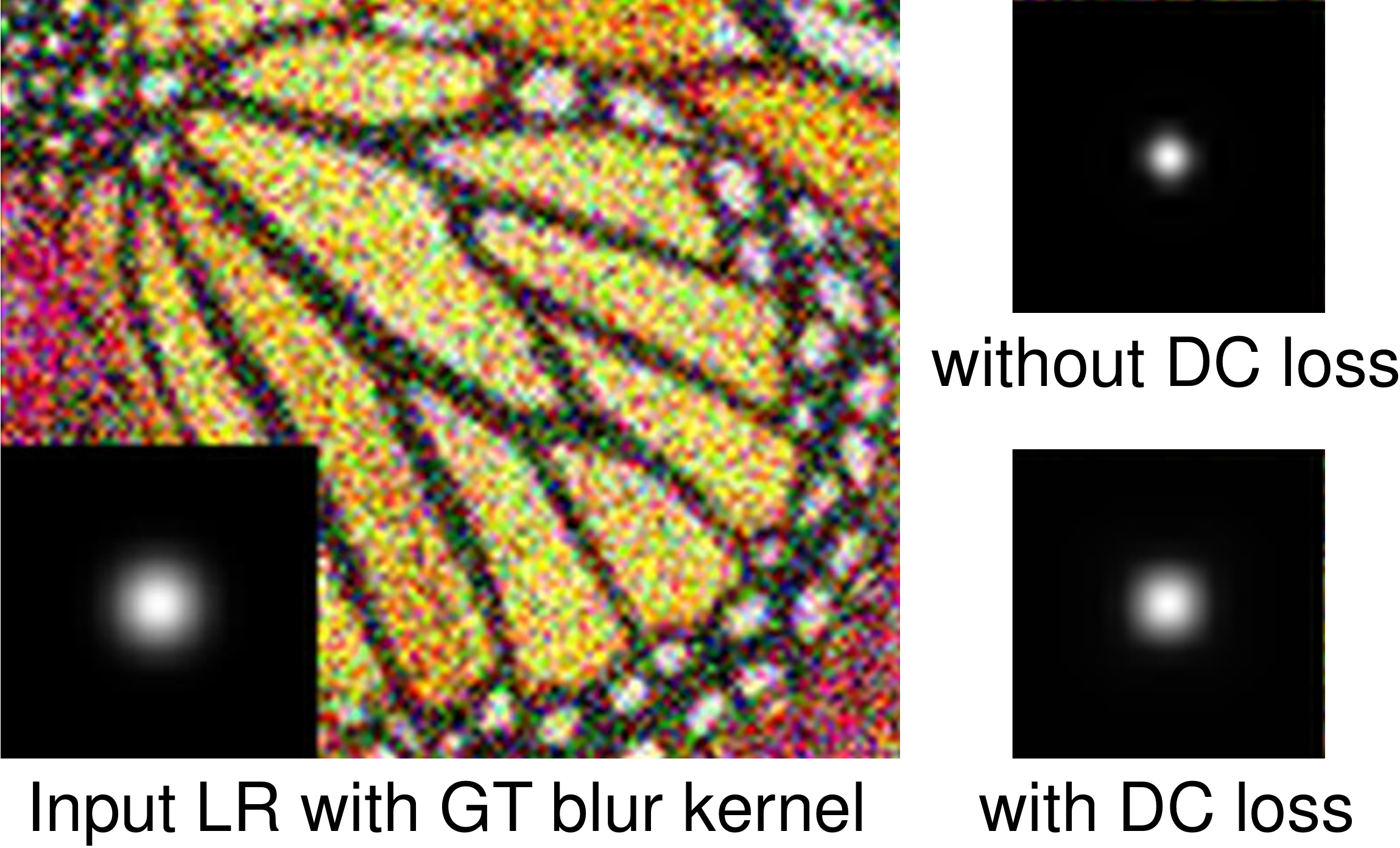}
  \caption{Ablation on degradation consistency loss ($\times4$, blur kernel width 1.3, noise level 50). [Best viewed in color]}
  \label{fig:ablation_dc}
\end{figure}

\textbf{Degradation Extractor}. In this part, we will verify the effectiveness of the proposed DR loss and DC loss. As shown in Table~\ref{tab:ablation_de}, without all losses in the degradation extractor, PSNR / SSIM performance are 24.87 / 0.7194. When adding DR loss to directly supervise degradation estimation, the performance increases to 24.92 / 0.7220. After applying DC loss on the LR image and degradation (blur kernel and noise map), the final performance is further increased to 24.98 / 0.7256. Such an ablation demonstrates the effectiveness of the losses in our degradation extractor. In addition, Figure~\ref{fig:ablation_dc} shows the blur kernel estimation results with and without the degradation consistency loss. With DC loss, the estimated blur kernel is more accurate.

\begin{table}[t]
\caption{Ablation on degradation reconstruction (DR) loss and degradation consistency (DC) loss. (LR) means applying DC loss on the LR image while (KN) indicates applying DC loss on the blur kernel and noise map ($\times$4, blur kernel 1.3, noise level 50).}
\vspace{1mm}
\label{tab:ablation_de}
\centering
\scalebox{0.95}{
\begin{tabular}{c|c|c|c}
\hline
DR loss      &DC loss (LR) &DC loss (KN)  &PSNR / SSIM \\
\hline
\hline
             &             &              &24.87 / .7194 \\
\hline
\checkmark   &             &              &24.92 / .7220 \\
\hline
\checkmark   &\checkmark   &              &24.95 / .7239 \\
\hline
\checkmark   &\checkmark   &\checkmark    &\textbf{24.98} / \textbf{.7256} \\
\hline
\end{tabular}
}
\end{table}

\textbf{Meta-restoration Modules}.
We also conduct ablation experiments on the two types of meta-restoration modules, MNM and MBM. The ablation results can be viewed in Table~\ref{tab:ablation_mrm}. For meta-denoise module, we first verify that the noise map is superior to the noise level number $\sigma$. We run a comparison model in which the degradation extractor predicts the noise level $\sigma$ and this number will be spatially repeated as the ``noise map'' to be used in MNM. From the table, we can see that utilizing the noise map will bring about 0.1 PSNR performance improvement, which demonstrates the superiority of the noise map. For meta-deblur module, when we add MBM into the model, the performance will be further increased. Such ablation experiments verifies the effectiveness of our two types of meta-restoration modules.

\begin{table}
\caption{Ablation on meta-restoration modules. -$\sigma$ and -map indicate adopting noise level $\sigma$ or noise map to express noise degradation, respectively ($\times2$, blur kernel 2.6, noise level 15).}
\vspace{1mm}
\label{tab:ablation_mrm}
\centering
\begin{tabular}{c|c|c|c}
\hline
MNM-$\sigma$  &MNM-map     &MBM         &PSNR / SSIM \\
\hline
\hline
\checkmark    &            &            &28.61 / .8119 \\
\hline
\checkmark    &            &\checkmark  &28.70 / .8148 \\
\hline
              &\checkmark  &            &28.72 / .8156 \\
\hline
              &\checkmark  &\checkmark  &\textbf{28.81} / \textbf{.8171} \\
\hline
\end{tabular}
\end{table}

\textbf{The number of MBM.} We also analyze the model performance with different numbers of MBMs. As shown in Table~\ref{tab:ablation_mbm_number}, we conduct experiments on one, two and three MBMs. After one MBM equipped, adding more MBMs can not bring obvious performance increase. This is because multiple MBMs is equal to one MBM by its linear nature. Considering additional memory cost, our final model contains only one MBM as the default setting.

\begin{table}
\caption{Ablation on different numbers of MBM. ($\times 3$, blur kernel 1.3, noise level 50).}
\vspace{1mm}
\label{tab:ablation_mbm_number}
\centering
\begin{tabular}{c|c|c|c}
\hline
 1 MBM  &2 MBM  &3 MBM  &PSNR / SSIM \\
\hline
\hline
\checkmark    &            &            &25.96 / 0.7533 \\
\hline
              &\checkmark  &            &\textbf{25.97} / \textbf{0.7545}  \\
\hline
              &            &\checkmark  &25.95 / 0.7537  \\
\hline
\end{tabular}
\end{table}

\section{Conclusion}
In this paper, we propose a \textbf{D}egradation-guided \textbf{M}eta-restoration network for blind \textbf{S}uper-\textbf{R}esolution (DMSR) which aims to restore a real-world LR image to an HR image. DMSR consists of two types of meta-restoration modules, and a degradation extractor optimized by a tailored degradation consistency loss. The blur and noise degradations can be estimated online from the extractor and further guide the meta-restoration modules to generate restoration parameters. Therefore, our DMSR model can handle real-world LR images with arbitrary degradations. Extensive experiments on benchmarks and real cases demonstrate the real-world application ability of our model. Yet, we can still observe failure cases in some complicatedly degraded image. We will consider more types of degradations as our future work to enhance the practical application ability of our model, such as motion blur, salt-pepper noise or jpeg block artifacts.

\clearpage
{\small
\bibliographystyle{ieee_fullname}
\bibliography{dmsr_arxiv}
}

\clearpage
\appendix
\renewcommand{\thesection}{\Alph{section}}
\renewcommand\thefigure{\thesection.\arabic{figure}}
\renewcommand\thetable{\thesection.\arabic{table}}

\setcounter{section}{0}
\section*{Supplementary}
In this supplementary material, Section~\ref{sec:time_size} describes the comparison of the running time and the parameter number. Section~\ref{sec:network_structure} illustrates the network structure details of our proposed DMSR. In Section~\ref{sec:degradation_window}, we show the degradation parameter window which are used for parameter chosen in non-blind SR methods. Finally, more visual results will be shown in Section~\ref{sec:visual}.

\section{Running Time and Parameter Number}\label{sec:time_size}
In this section, the inference time and the parameter number of DMSR will be discussed. Our DMSR model is compared with the SOTA blind SR approaches which are adopted in our paper. These approaches include ZSSR~\cite{shocher2018zero}, IKC~\cite{gu2019blind} and SFM~\cite{elhelou2020stochastic}. Because we choose the blind SR SOTA method IKC as the backbone of SFM, so their inference time and parameter number are the same. For running time, all the models are run on a single RTX 2080 GPU with an input LR image of the size $128 \times 128$. Table~\ref{tab:time_size} shows the results with scale factor $\times$2, in which ZSSR has the smallest parameter number but its inference time is extremely high. This is because for every input LR image, ZSSR needs to first train on the LR image before generating SR result during inference, which is a restriction for practical application. IKC and SFM spend more time than our DMSR model in inference due to the iterative strategy. Our DMSR model can achieve significantly better performance than SOTA methods with less parameter number and inference time.

\section{Details of Network Structure}\label{sec:network_structure}
Our DMSR model contains a degradation extractor and an SR network. The structure of the degradation extractor is illustrated in Table~\ref{tab:structure_de}. A convolution layer and two residual blocks are adopted to extract image features. In the noise branch, an additional convolution layer predicts the noise map. In the blur branch, two convolution layers, a global pooling layer and a softmax layer are adopted sequentially to estimate the blur kernel with the size of $15 \times 15$.

Our SR network is composed of two types of meta-restoration modules (MNM and MBM), a deep feature learning and up-sampling part. The structure of the SR network is shown in Table~\ref{tab:structure_sr}. We divide the SR network into three parts, in which the last two columns are the layers for MNM and MBM.

\begin{table}
\newcommand{\tabincell}[2]{\begin{tabular}{@{}#1@{}}#2\end{tabular}}
\caption{Parameter number and inference time of different blind SR approaches. All approaches have the scale factor of $\times$2 with input LR image of the size $128 \times 128$.}
\label{tab:time_size}
\begin{center}
\scalebox{0.95}{
\begin{tabular}{|c|c|c|c|c|}
\hline
\multicolumn{2}{|c}{\multirow{2}{*}{Approach}} &\multicolumn{2}{|c|}{Parameter Number} &\multirow{2}{*}{Time}  \\
\cline{3-4}
\multicolumn{2}{|c|}{} &Part &Total & \\
\hline
ZSSR~\cite{shocher2018zero} & / &/ &0.22M &39.758s\\
\hline
\multirow{3}{*}{\tabincell{c}{IKC~\cite{gu2019blind} / \\SFM~\cite{elhelou2020stochastic}}} &Predictor   &0.43M &\multirow{3}{*}{8.90M} &\multirow{3}{*}{0.617s}\\
\cline{2-3}
                                        &Corrector   &0.65M & &\\
\cline{2-3}
                                        &SFTMD       &7.82M & &\\
\hline
\multirow{2}{*}{DMSR}                   &DE          &0.48M &\multirow{2}{*}{8.37M} &\multirow{2}{*}{0.595s}\\
\cline{2-3}
                                        &SR network  &7.89M &      &\\
\hline
\end{tabular}
}
\end{center}
\end{table}

\begin{table}
\caption{The network structure of the degradation extractor. Conv($C_{in}$, $C_{out}$) indicates the convolution layer with $C_{in}$ input channels and $C_{out}$ output channels. ResBlock($C$) represent a residual block with channel number $C$.}
\label{tab:structure_de}
\begin{center}
\begin{tabular}{|c|c|c|}
\hline
Id &\multicolumn{2}{c|}{Layer name(s)}\\
\hline
\hline
0 &\multicolumn{2}{c|}{Conv(3, 64), ReLU}\\
\hline
1 &\multicolumn{2}{c|}{ResBlock(64)} \\
\hline
2 &\multicolumn{2}{c|}{ResBlock(64)} \\
\hline
3 &Conv(64, 128), ReLU &\multirow{4}{*}{Conv(64,3)} \\
\cline{1-2}
4 &Conv(128, 225), ReLU &  \\
\cline{1-2}
5 &GlobalPool       &  \\
\cline{1-2}
6 &Softmax   & \\
\hline
\end{tabular}
\end{center}
\end{table}

\begin{table*}[t]
\caption{The network structure of the SR network in DMSR. $I^{LR}$ indicates the input degraded LR image. $N_{est}$ and $K_{est}$ represent the estimated noise map ($H \times W \times 3$) and blur kernel ($15 \times 15$) from the degradation extractor. Concat() means the concatenation operation. Conv($C_{in}$, $C_{out}$) is the convolution layer with $C_{in}$ input channels and $C_{out}$ output channels, while FC($U_{in}$, $U_{out}$) is the fully connected layer with $U_{in}$ input units and $U_{out}$ output units. DynamicConv($k_d$) indicates the dynamic convolution with dynamic kernel $k_d$. The up-sampling scale factor is denoted as $c$. Five residual groups which are proposed in RCAN~\cite{zhang2018image} are adopted in the network.}
\label{tab:structure_sr}
\begin{center}
\begin{tabular}{|c|c|c|c|c|c|c|}
\hline
Id  &Layer name(s) &Id &Layer name(s) for MNM &Id &Layer name(s) for MBM &Output size\\
\hline
\hline
\multirow{2}{*}{\ } &\multirow{2}{*}{\ }        &1-0                   &Concat($I^{LR} \| N_{est}$)  &\multirow{6}{*}{\ }       &                         &\multirow{6}{*}{\ } $H \times W \times 6$\\
\cline{3-4} \cline{7-7}
                    &                           &1-1                   &Conv(6, 64)                   &                          &                         & $H \times W \times 64$\\
\cline{1-4} \cline{7-7}
0-0                 &Conv(64, 64)(\#1-1)        &\multirow{10}{*}{\ }  &\multirow{10}{*}{\ }          &                          &                         & $H \times W \times 64$\\
\cline{1-2} \cline{7-7}
0-1                 &Residual Group $\times 5$ &                      &                              &                          &                         & $H \times W \times 64$\\
\cline{1-2} \cline{7-7}
0-2                 &Conv, PixelShuffle, ReLU   &                      &                              &                          &                         & $cH \times cW \times 64$\\
\cline{1-2} \cline{7-7} 
0-3                 &Conv(64, 3)                &                      &                              &                          &                         & $cH \times cW \times 3$\\
\cline{1-2} \cline{5-7}
\multirow{4}{*}{\ } &\multirow{4}{*}{\ }        &                      &                              &2-0                       &FC(225, 15)($K_{est}$)   &$1 \times 1 \times 15$\\
\cline{5-7}
                    &                           &                      &                              &2-1                       &Repeat Spatially         &$cH \times cW \times 15$\\
\cline{5-7}
                    &                           &                      &                              &2-2                       &Concat(\#0-3 $\|$ \#2-1) &$cH \times cW \times 18$\\
\cline{5-7}
                    &                           &                      &                              &2-3                       &Conv(18, 225)            & $cH \times cW \times 225$\\
\cline{1-2} \cline{5-7}
0-4                 &DynamicConv(\#2-3)         &                      &                              &                          &                         &$cH \times cW \times 3$ \\
\cline{1-2} \cline{7-7}
0-5                 &Conv(3, 3)                 &                      &                              &                          &                         &$cH \times cW \times 3$\\
\hline 
\end{tabular}
\end{center}
\end{table*}

\section{Degradation Parameter Window}\label{sec:degradation_window}
\begin{figure}
\centering
  \includegraphics[width=\linewidth]{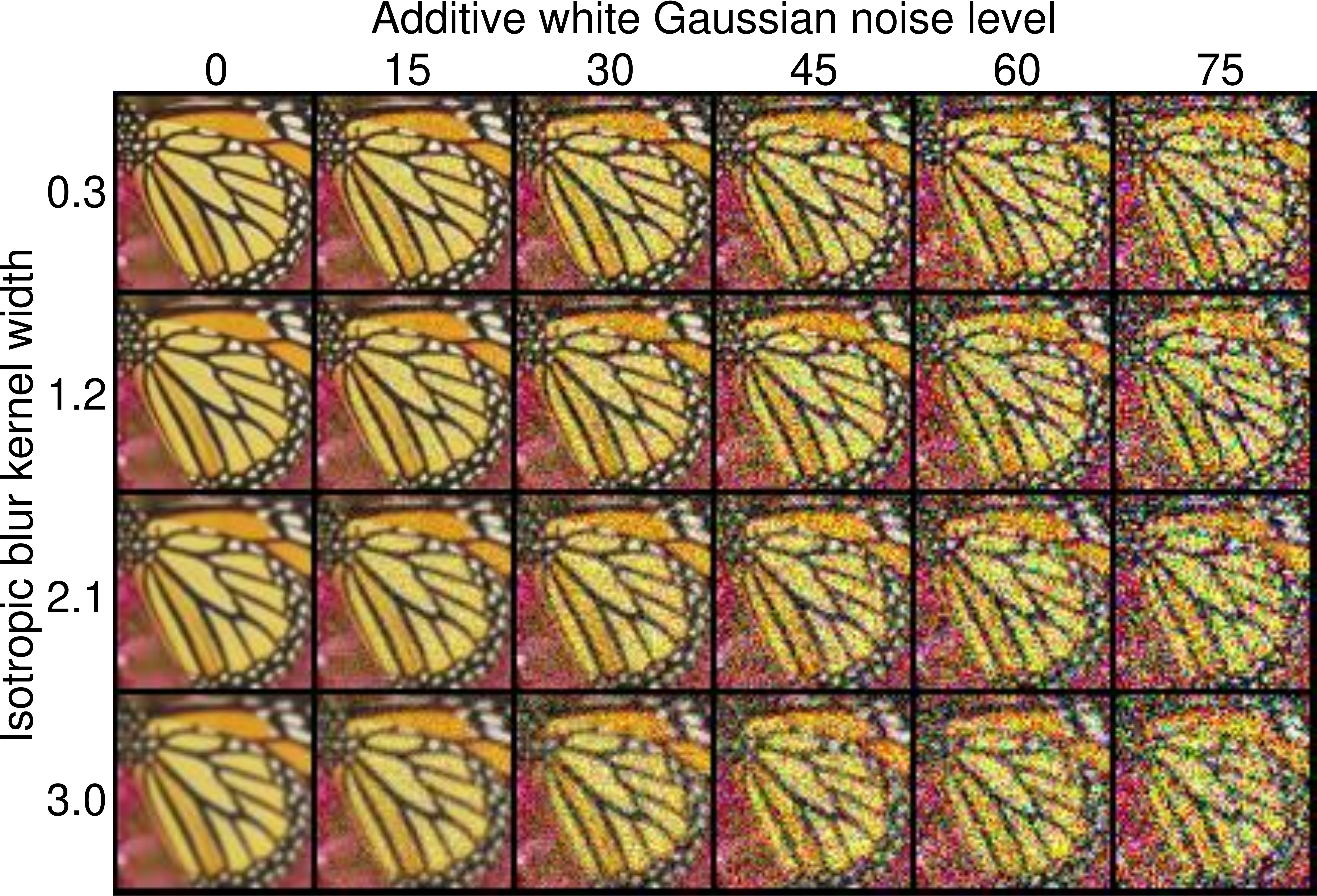}%
  \caption{Degradation parameter window for the degradation parameter choice in non-blind SR methods.}
  \label{fig:degradation_window}
\end{figure}

In this part, we show the degradation parameter window for the degradation parameter choices in non-blind SR methods. As shown in Figure~\ref{fig:degradation_window}, during inference, we choose the most similar degraded image in the window and adopt its parameters as the input for non-blind SR methods.

\begin{figure*}
\centering
  \begin{tabular}{p{4mm}p{21mm}p{20mm}p{18mm}p{19mm}p{19mm}p{24mm}p{15mm}p{10mm}}
        \ 
		&\scriptsize{ZSSR~\cite{shocher2018zero}}
		&\scriptsize{IKC~\cite{gu2019blind}}
		&\scriptsize{SFM~\cite{elhelou2020stochastic}}
		&\scriptsize{IKC~\cite{gu2019blind}-vn}
		&\scriptsize{SFM~\cite{elhelou2020stochastic}-vn}
		&\scriptsize{DMSR (ours)}
		&\scriptsize{GT}
		&\ \ \ 
  \end{tabular}\\
  \includegraphics[width=0.95\linewidth]{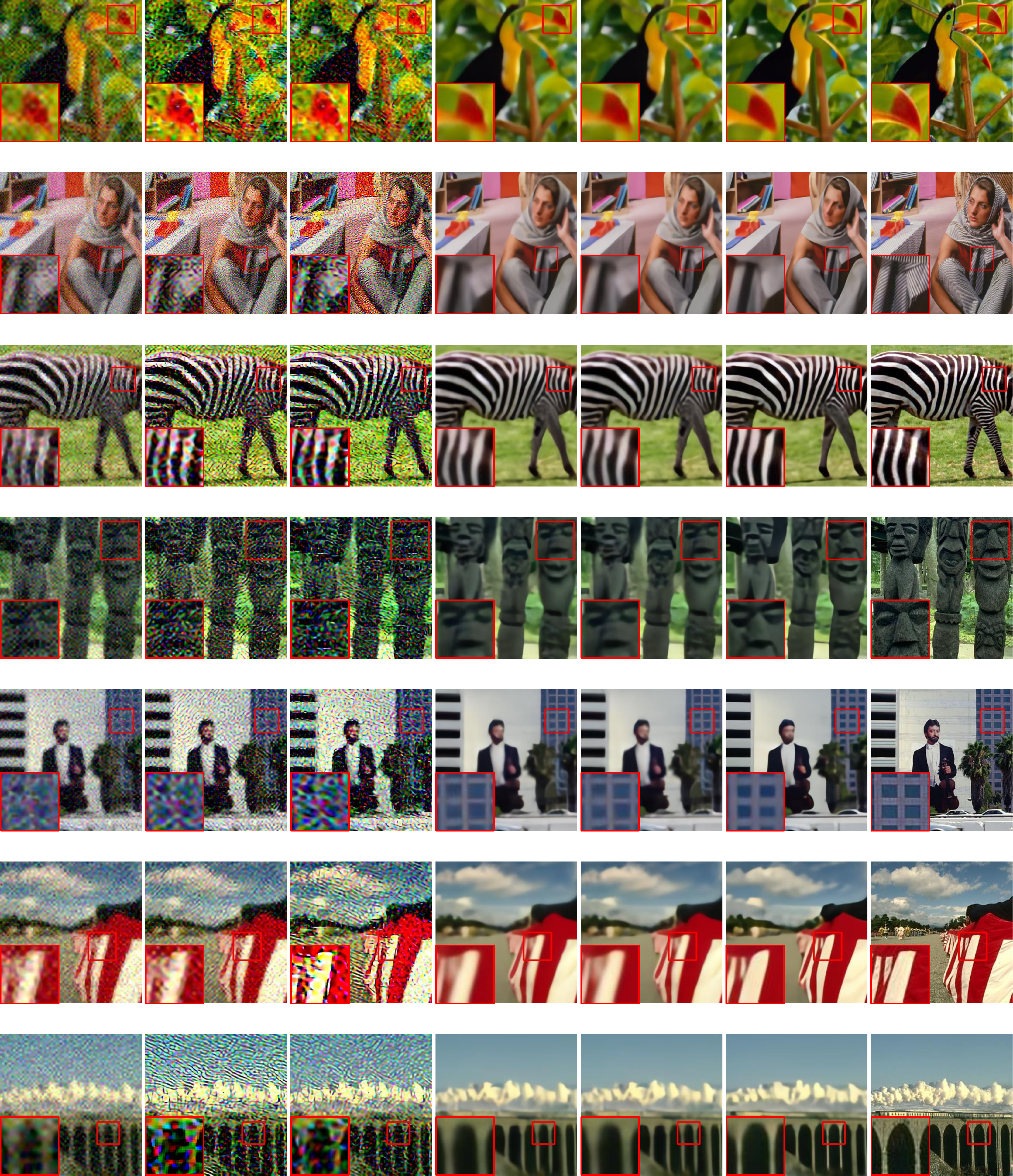}%
  \caption{More visual comparison among different blind SR methods on Set5~\cite{bevilacqua2012low}, Set14~\cite{zeyde2010single} and B100~\cite{martin2001database} datasets. Degradation: isotropic Gaussian blur kernel width 2.6, additive white Gaussian noise 15 and scale factor $\times$4. [Best viewed in color]}
  \label{fig:visual_blindsr_supp}
\end{figure*}

\section{More Visual Results}\label{sec:visual}
\begin{figure*}[t]
\centering
  \includegraphics[width=\linewidth]{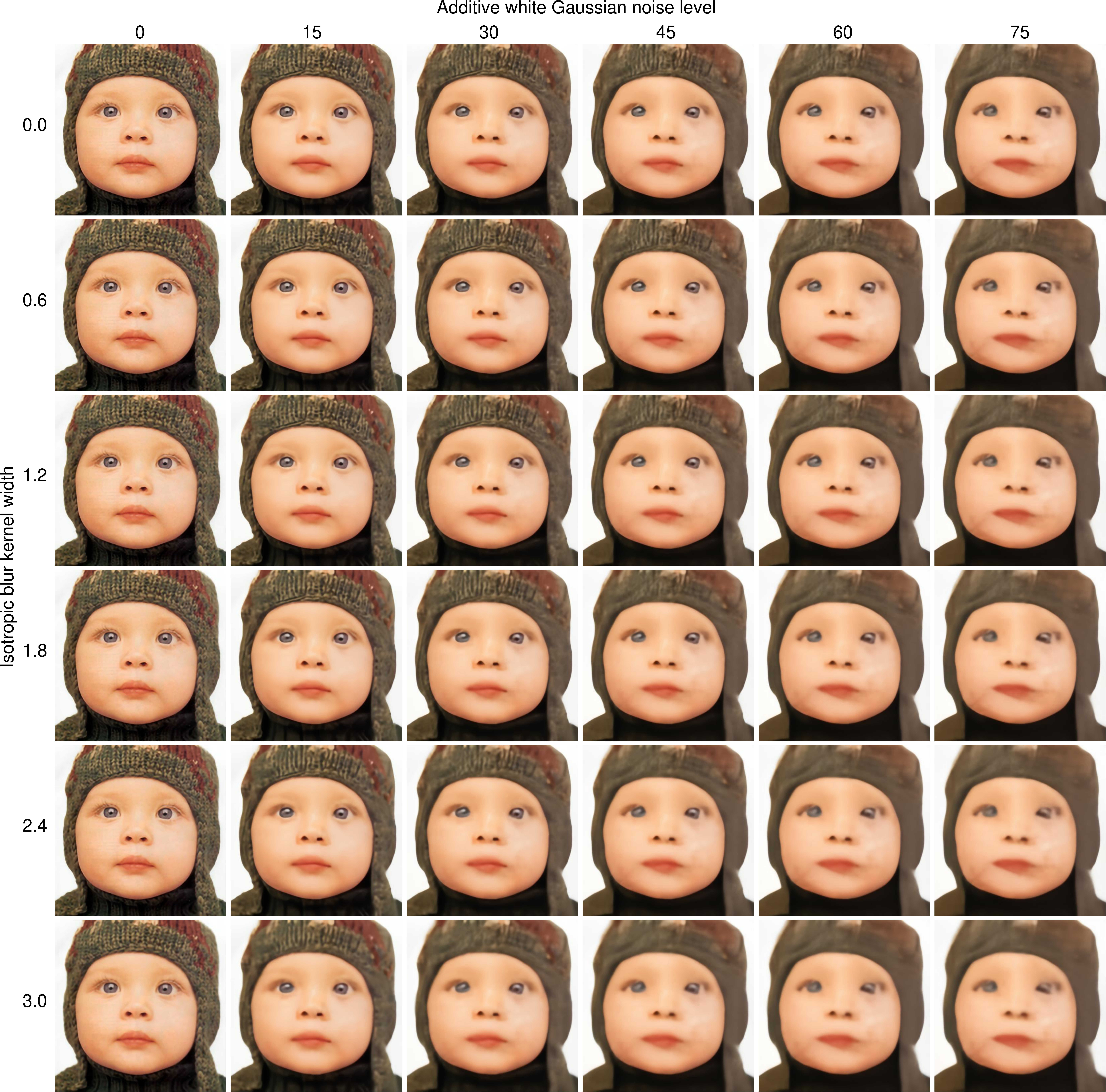}
  \caption{Our DMSR model results on different degradations with scale factor $\times$4.}
  \label{fig:results_different_degradations}
\end{figure*}

In this section, we show more blind SR results on the common-used datasets, Set5~\cite{bevilacqua2012low}, Set14~\cite{zeyde2010single} and B100~\cite{martin2001database}, as shown in Figure~\ref{fig:visual_blindsr_supp}. These results further demonstrate the effectiveness of our DMSR model over SOTA models. In addition, to verify the robustness of our proposed DMSR model, we show the SR results on different degradation situations in Figure~\ref{fig:results_different_degradations}. We choose a wide range of blur kernel widths and noise levels as degradations with scale factor $\times$4. The results demonstrate that our model can handle a wide range of degradations in real-world scenarios.

\begin{figure*}[t]
\centering
  \includegraphics[width=\linewidth]{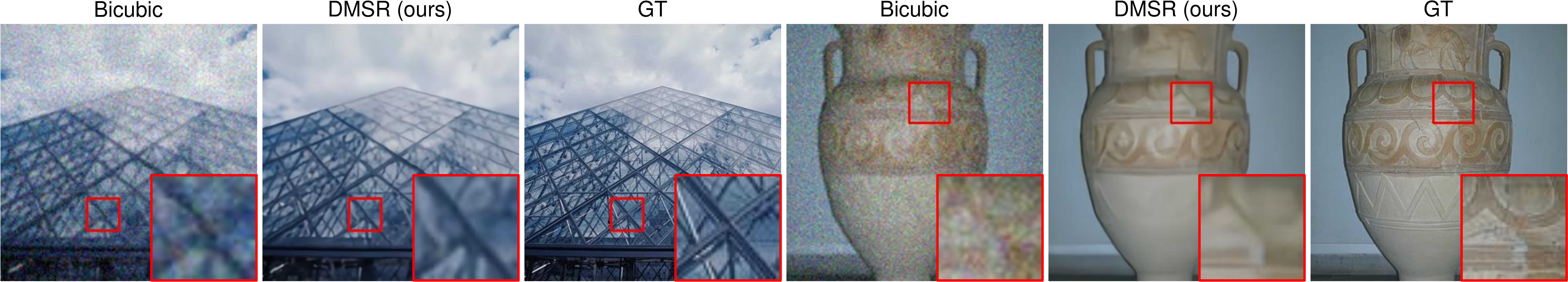}
  \caption{Examples of some failure cases. The degradation parameters are $\times 2$, Gaussian blur kernel width 1.3, and additive white Gaussian noise level 15. Our model fails to generate accurate textures due to the influence of the severe degradation.}
  \label{fig:failure_cases}
\end{figure*}

We also observe some failure cases during our experiments and we show them in Figure~\ref{fig:failure_cases}. Due to the severe noise and blur degradation, our DMSR model fails to generate accurate textures in some special cases. We will further study more effective models to solve such problems.

\end{document}